# Artificial Intelligence in Minimally Invasive Interventional Treatment

Daniel Ruijters[1,2]

[1] Philips Healthcare, Image Guided Therapy Systems Innovation, Veenpluis 6, 5684PC, the Netherlands
[2] Technische Universiteit Eindhoven, Dept. Electrical Engineering, Den Dolech 2, 5612AZ Eindhoven

*Minimally invasive image guided treatment procedures often employ advanced image processing algorithms. The recent developments of artificial intelligence algorithms harbor potential to further enhance this domain. In this article we explore several application areas within the minimally invasive treatment space and discuss the deployment of artificial intelligence within these areas.*

## Introduction

Minimally invasive treatment, including procedures such as coronary artery stenting, trans-arterial chemo embolization in the liver, or vertebroplasty, relies quite heavily on sophisticated image processing algorithms, often employed in (near) real-time. In recent years artificial intelligence (AI) algorithms, such as convolutional networks, are seeing increasing popularity, fueled by the availability of large datasets in social media networks and the development of standard platforms. These developments have also proven to provide benefits to medical applications, often surpassing the performance of more traditional algorithmic approaches [1]. Particularly, the plasticity of AI algorithms, their ability of adapting to new input and their generalizing properties are of advantage in a minimally invasive treatment context. In this article, we explore a number of potential applications of such AI algorithms in minimally invasive procedures.

## Device detection

Typically, in minimally invasive procedures the in-body device can only be navigated and monitored through external imaging. Suitable imaging modalities are ultrasound, interventional x-ray, and real-time CT and MR. AI can be employed to detect and locate interventional devices, such as intra-vascular devices, and other percutaneous devices. Intra-vascular devices comprise catheters and guidewires [2], stents, intra-vascular valves, etc. Other percutaneous devices entail needles [3], scalpels, etc. AI algorithms are particularly suited for detecting and segmenting devices since they are trained by a suitable set of examples [2,4,5]. This implies that the training set can contain a variety of devices with different visual properties, and it can be easily extended with new devices.

## Multi-modal image fusion

Integrating data from various imaging modalities can aid the interventional treatment procedure. E.g., pre-interventional planning conducted on diagnostic images can be utilized during the procedure [6], see Figure 1. The spatial registration of the pre-interventional and peri-interventional images can then bring the pre-interventional planning into the coordinate space of the interventional equipment. This allows to overlay the planning, such as a needle path, on the live images containing the interventional devices. Also, multiple complementary imaging modalities, such as e.g. ultrasound and x-ray, can be combined to create richer more informative data [7]. The combination of the images can show interfaces between tissues and objects that can only be visualized by a different imaging modality.

The spatial co-registration process can be conducted based on explicit markers and other external knowledge, on image content alone, or a combination of those. The resulting spatial mapping can be rigid, affine, elastic, or other deformable, depending on the clinical application. E.g., for intra-cranial applications a rigid registration is often sufficient, while registering pre- and intra-interventional abdominal images may require elastic registration to account for respiratory motion, etc. [8].

AI based approaches may play a role in establishing the spatial co-registration mapping either by detecting explicit landmarks and/or identifying landmark features in images, or by integrally addressing the registration task [9,10].

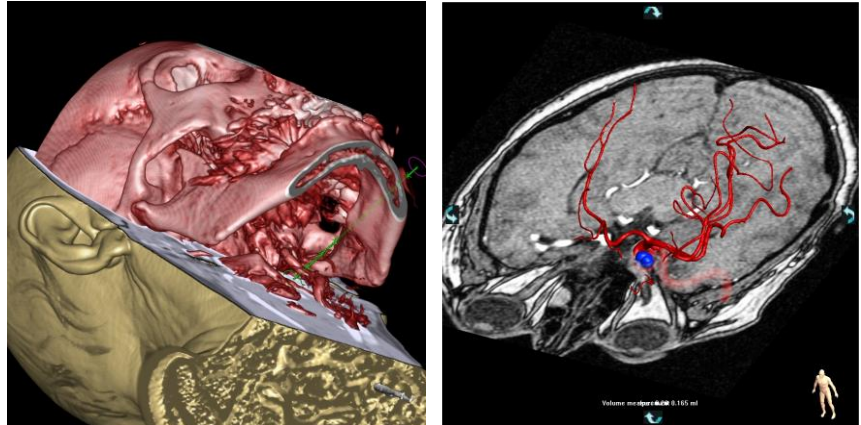

*Figure 1: Examples of spatially registered multi-modal datasets*

## Functional imaging

Functional imaging has as purpose to characterize the functioning of biological processes, rather than visualizing the anatomy (though it is typically combined with anatomical imaging in order to localize the functional aspects). Examples of the functions that are imaged peri-interventionally are blood flow in vessels and aneurysms [11,12], blood perfusion of the parenchymal tissue, valve motion, etc. Functional imaging is typically based on intensive processing of raw measurements. For e.g. blood flow vector fields through digital subtraction angiography (see Figure 2), the motion of contrast through the vascular structures is followed in the consecutive frames, while for valve motion the valve leaflets are segmented and followed over time. These segmentations and motion of carrier substances are very well suited for AI approaches, such as convolutional networks [2,4].

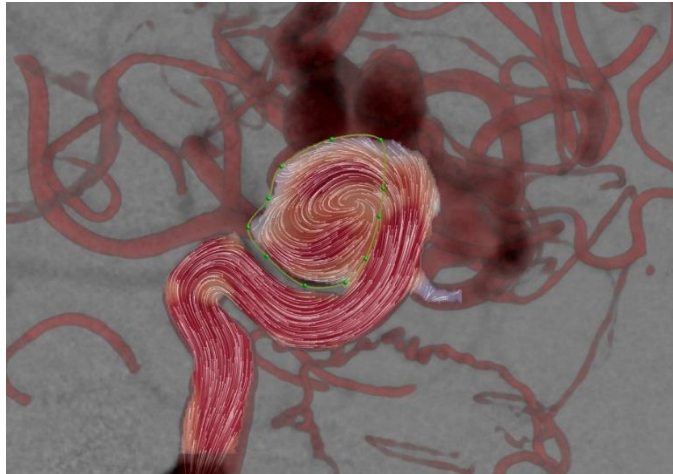

*Figure 2: Example of intra-vascular flow imaging using digitally subtracted angiography imaging.*

## Stereoscopic displays

Finally, visualization in an interventional context needs to be as easy to interpret as possible, because many stimuli are fighting for attention in the intervention room. For 3D data stereoscopic visualization aids in making the interpretation as intuitive as possible. Autostereoscopic displays [13] are very suited for this task, as they do not require that the users wear dedicated goggles or other gear. AI algorithms can be employed to calculate missing viewing angles through free viewpoint interpolation [14], or even add per pixel depth to a 2D image [15].

## Conclusions

In this article, we have explored several application areas in the minimally invasive interventional domain, with respect to the deployment of artificial intelligence applications. We have zoomed in on device detection, multi-modal image fusion, functional imaging and stereoscopic displays, and discussed where and how AI can be exploited to further advance these application areas.